\documentclass[runningheads]{llncs}

\usepackage[T1]{fontenc}
\usepackage{amsmath}
\usepackage{amssymb}
\usepackage{amsfonts}
\usepackage{graphicx}
\usepackage{booktabs}
\usepackage{tabularx}
\usepackage{array}
\usepackage{algorithm}
\usepackage{algorithmic}
\usepackage{placeins}
\usepackage{marvosym}

\newcolumntype{C}{>{\centering\arraybackslash}X}

\begin{document}

\title{FROD: Feature Matching Residual Denoising Oracle Bone Decipher}
\titlerunning{FROD}

\author{Yanbin Hou\inst{1} \and Biao Xiong\inst{1} \and Guojun Xu\inst{1} \and Jianwen Xiang\inst{1} \and Cheng Tan\inst{1} \and Yanchao Yang\inst{1} \and Junwei Zhou\inst{1}\textsuperscript{(\Letter)}}
\authorrunning{Y. Hou et al.}

\institute{School of Artificial Intelligence, Wuhan University of Technology, Wuhan 430070, China\\
\email{houyanbin@whut.edu.cn, b.xiong@whut.edu.cn, guojunxu@whut.edu.cn, jwxiang@whut.edu.cn, cheng\_tan@whut.edu.cn, yangyc@whut.edu.cn, junweizhou@msn.com}}

\maketitle

\begin{abstract}
Oracle bone script (OBS), one of the earliest Chinese writing systems, plays an important role in the study of Chinese etymology. Traditional decipherment relies heavily on domain experts who analyze characters through semantic context and structural evolution. To assist this labor-intensive process, we formulate OBS decipherment assistance as a cross-era image translation task and propose FROD (Feature Matching Residual Denoising Oracle Bone Decipher). Although many OBS characters differ substantially from their modern counterparts, they often preserve local topological invariants at the radical level. During training, FROD leverages fast feature matching to provide gated segmentation supervision: paired samples with sufficient matches are processed patch-wise to align fine-grained radicals, whereas low-similarity pairs are trained holistically to avoid mismatched artifacts. In addition, a Residual Denoising Diffusion Model (RDDM) jointly estimates noise and residual signals, thereby reducing the positional drift and stroke disorder commonly observed in standard diffusion models. Finally, a multi-stage font stylization refinement network refines the generated images by eliminating edge noise and stabilizing stroke structures. On our augmented character-disjoint dataset, FROD achieves higher Top-1 recognition accuracy than the evaluated baselines, with a 3.8\% absolute gain over OBSD.
\end{abstract}

\section{Introduction}

Oracle bone script is one of the earliest forms of pictographic writing and was used during the Shang Dynasty. The study of oracle bone script (OBS) is fundamental to research on Chinese etymology and ancient history. However, among the approximately 4,500 unique characters discovered so far, only about 1,500 have been reliably deciphered. The decipherment process remains labor-intensive, requiring substantial domain knowledge and associative reasoning to map ancient symbols to their modern counterparts.

Although end-to-end decipherment ultimately depends on contextual and linguistic evidence, generating visually recognizable modern Chinese character candidates from OBS rubbings can greatly accelerate the research workflow. Recent studies have therefore formulated this problem as an image-to-image translation task. In particular, OBSD~\cite{guan2024deciphering} employs a diffusion model with Local Structural Sampling (LSS) to align fine-grained patches between OBS and modern characters. However, OBS characters often exhibit substantial structural variation. Forcing highly abstract or weakly aligned pairs into naive patch-wise mappings can lead to severe positional drift, mismatched artifacts, and stroke disorder. In such cases, blind segmentation becomes unreliable because explicit structural correspondences are weak.

To address these challenges, we propose FROD, an image translation framework for OBS decipherment assistance. Compared with OBSD, our approach offers three key improvements. First, rather than blindly segmenting all training pairs, we use the fast feature-matching model LightGlue~\cite{lindenberger2023lightglue} to provide gated segmentation supervision. During training, paired OBS and modern characters with sufficient topological overlap, such as shared radicals, are processed patch-wise to align local features, whereas highly variable pairs are trained holistically to preserve global structure. Second, we adopt the Residual Denoising Diffusion Model (RDDM)~\cite{liu2024residual} in place of a standard diffusion model. By modeling target residuals and generated noise as separate components, RDDM alleviates the positional drift and structural bias that often arise in conventional noise-prediction frameworks. Third, we apply a multi-stage font stylization refinement network~\cite{fu2024generate} to refine the synthesized outputs, suppress edge noise, and produce standardized modern glyphs.

Our main contributions are as follows:
\begin{itemize}
\item We introduce a feature-matching-gated training strategy for OBS image translation. By assigning paired training samples to patch-wise or holistic supervision according to structural similarity, our method avoids detrimental misalignments while capturing local radical correspondences when appropriate.
\item We adapt a Residual Denoising Diffusion Model to the OBS domain, which explicitly models residual signals to better capture stroke positional patterns and preserve structural fidelity compared to standard diffusion baselines.
\item We integrate a font stylization refinement module to refine the predicted glyphs. Extensive experiments show that the full pipeline achieves the best OCR accuracy among the evaluated baselines on our augmented character-disjoint dataset and improves Top-1 OCR accuracy over OBSD.
\end{itemize}

By producing reliable modern character candidates, FROD helps bridge computer vision and archaeology and provides a practical assistive tool for OBS decipherment.

\section{Related Work}

OBS research has advanced substantially with the release of several digitized datasets. Representative collections include Oracle-20K~\cite{guo2015building}, OBC306~\cite{huang2019obc306}, and HWOBC~\cite{li2020hwobc}. However, these early datasets focus primarily on isolated character recognition and do not provide structural evolutionary mappings between OBS and modern Chinese characters. More recent resources, such as EVOBC~\cite{guan2024open}, HUST-OBC~\cite{wang2024open}, and OBC-V~\cite{zhou2025oracle}, address this limitation by providing modern-character correspondences and expanded class coverage, thereby laying the foundation for translation-based approaches.

For image-to-image translation, conditional Generative Adversarial Networks (GANs) such as Pix2Pix~\cite{isola2017image}, CycleGAN~\cite{zhu2017cyclegan}, and DRIT++~\cite{lee2020dritpp}, as well as diffusion models such as Palette~\cite{saharia2022palette} and BBDM~\cite{li2023bbdm}, have shown strong performance. However, they often struggle to align highly deformed cross-era topologies without explicit structural guidance.

Within OBS studies, most prior work has focused on character recognition using computer vision or natural language processing techniques, whereas AI-assisted interpretation of undeciphered characters remains limited. Zhang et al.~\cite{zhang2021deciphering} proposed a case-based reasoning method that retrieves structurally similar cases from adjacent writing systems to assist expert decipherment. Chang et al.~\cite{chang2022sundial} introduced a cascaded GAN framework that models intermediate stages of Chinese-character evolution from OBS to modern forms. Guan et al.~\cite{guan2024deciphering} introduced a diffusion-based approach for detailed OBS-modern character alignment, thereby providing clearer evolutionary links.

Nevertheless, existing methods often struggle to preserve fine-grained structural differences between OBS and modern character images, which leads to positional drift, missing strokes, and blurred outputs. In addition, few pipelines include explicit font standardization after generation, which further limits the legibility of translated characters for downstream recognition.

\section{Feature Matching and Segmentation}

Due to the highly variable structure of OBS characters, establishing valid correlations with modern Chinese character images is essential. During paired training, blindly applying patch-wise supervision to highly abstract OBS-modern pairs often produces disordered or noisy strokes because blind segmentation forces alignments where explicit structural correspondences do not exist.

To address this issue, we introduce a training-time gating mechanism driven by fast feature matching. The number of high-confidence matched keypoints serves as a quantitative criterion for determining whether a paired training sample should receive patch-wise local supervision or holistic supervision.

\subsection{Feature Matching and Gating Mechanism}

We adopt LightGlue (LG)~\cite{lindenberger2023lightglue} to extract and match keypoints due to its efficiency on low-complexity images like skeletonized OBS strokes. LG utilizes self- and cross-attention layers enhanced by Rotary Positional Embeddings to capture spatial context. Based on these contextualized features, a lightweight head predicts point correspondences. Let $\mathbf{x}_i^I$ denote the descriptor of the $i$-th keypoint from the modern character image and $\mathbf{x}_j^S$ denote the descriptor of the $j$-th keypoint from the paired OBS image, where $M$ and $N$ are the corresponding numbers of detected keypoints. The similarity score matrix $\mathbf{S} \in \mathbb{R}^{M \times N}$ is computed as
\begin{align}
S_{ij} = (\mathbf{W}_{match}\mathbf{x}_i^I)^\top (\mathbf{W}_{match}\mathbf{x}_j^S).
\end{align}
With point-wise matchability confidences $\beta_i^I, \beta_j^S \in [0, 1]$, the soft assignment matrix $\mathbf{O}$ is formulated as
\begin{align}
O_{ij} = \beta_i^I \cdot \beta_j^S \cdot \text{Softmax}_{k \in I}(S_{kj})_i \cdot \text{Softmax}_{k \in S}(S_{ik})_j.
\end{align}
A match is considered valid if it is mutually maximal and its score $O_{ij}$ strictly exceeds a confidence threshold $\tau$. The number of valid matches is not treated as a semantic equivalence measure; instead, it serves as a practical proxy for the reliability of local structural correspondences between paired glyph images.

\begin{figure}[t]
\centering
\includegraphics[width=0.5\textwidth]{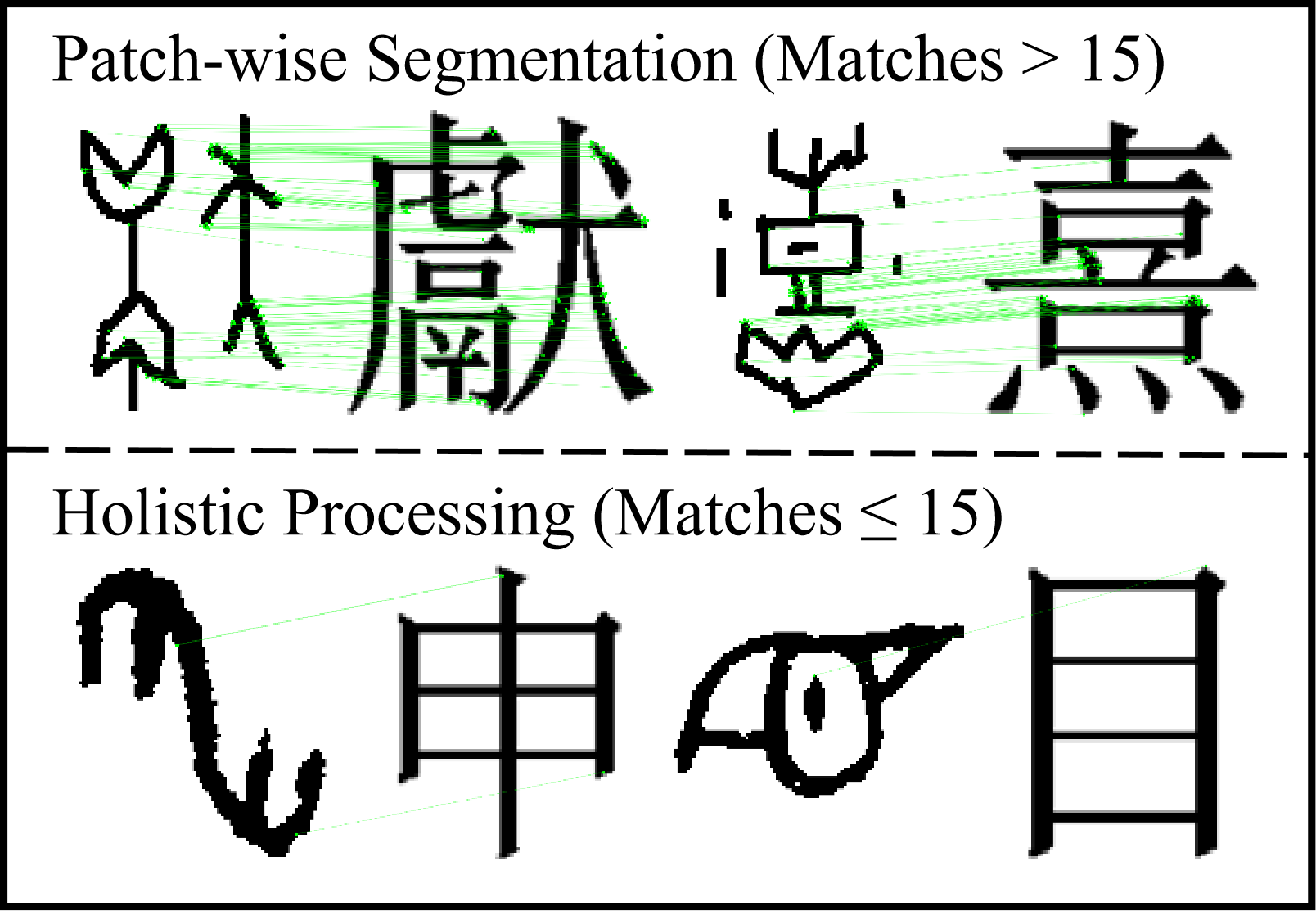}
\caption{Training-time gating mechanism for OBS (left) and modern characters (right). Top: dense keypoint matches ($n_{\mathrm{match}} > \gamma$) trigger patch-wise local segmentation for radical alignment. Bottom: sparse matches ($n_{\mathrm{match}} \leq \gamma$) lead to holistic processing to preserve global character topology, where $\gamma=15$.}
\label{fig:match}
\end{figure}

\textbf{Gating Threshold ($\gamma$):} During training, we count the total number of valid matches for each paired OBS-modern image sample. We set $\gamma = 15$ as a conservative empirical threshold to select pairs with sufficiently dense local correspondences for patch-wise supervision. If the match count exceeds this threshold, the pair exhibits sufficient structural correlation and is routed to the patch-wise segmentation module. Otherwise, it bypasses segmentation and is trained holistically to preserve global topology. As shown in Fig.~\ref{fig:match}, the upper examples exceed the threshold $\gamma$, triggering segmentation, while the lower ones do not.

\subsection{Image Segmentation Method}
\label{sec:segmentation}

For paired training samples that satisfy the gating condition, we adopt an LSS strategy. Specifically, the resized $100 \times 100$ input image $I \in \mathbb{R}^{100 \times 100 \times C}$ is divided into $D = 8$ overlapping local regions of size $p \times p$ (with $p = 64$) using a sliding window. To obtain exactly eight local regions while maintaining sufficient coverage of the character strokes, we use a grid sampling strategy with appropriate strides. In implementation, each local region is represented on a fixed $100 \times 100$ canvas before being fed into the diffusion network, while holistic samples also use the original $100 \times 100$ image. Thus, patch-wise and holistic training share the same network input resolution; $p$ only defines the local support region used for segmentation and blending.

These local patch representations, denoted as $I_0^{(d)}$ for the modern character and $\tilde{I}^{(d)}$ for the corresponding OBS character, are processed independently during training. A Gaussian blending mask $\mathbf{P}_d$ is generated for each patch position, which is later utilized during the reverse generation process to seamlessly blend overlapping regions and prevent visible seam artifacts.

\section{Residual Denoising Diffusion Method}

\begin{figure}[t]
\centering
\includegraphics[width=\textwidth]{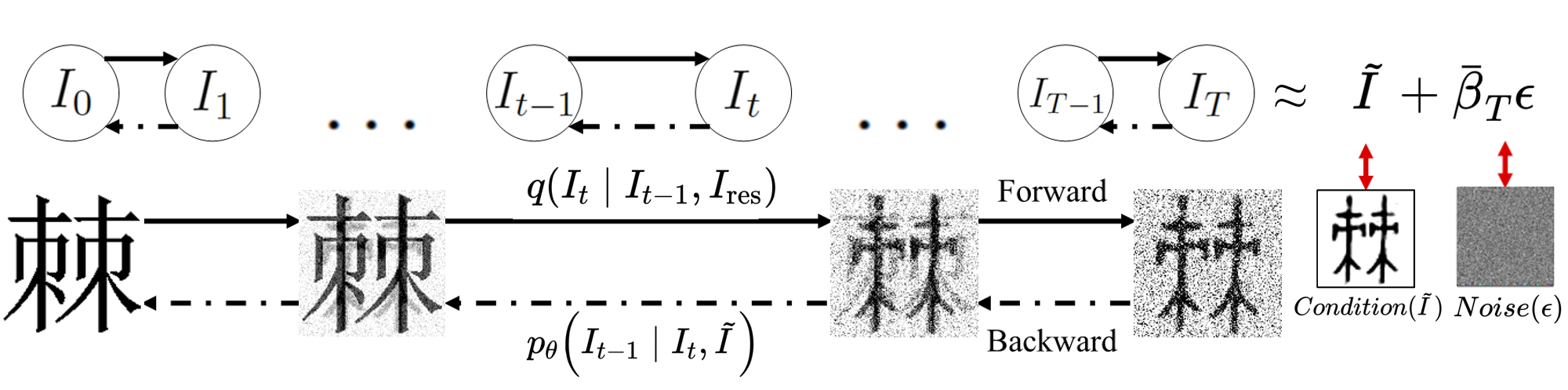}
\caption{Steps of the residual denoising diffusion model.}
\label{fig:model}
\end{figure}

Traditional conditional diffusion models typically estimate either the target image or the injected noise. Although methods such as I$^2$SB~\cite{liu2023i2sb} model bridges between paired image domains, they do not explicitly decouple deterministic domain residuals from stochastic noise. For OBS decipherment, where the structural gap between paired glyphs can be large, we adapt the RDDM framework of Liu et al.~\cite{liu2024residual} to OBS image translation. By explicitly treating the target residual and generated noise as independent components, this dual-estimation framework preserves the OBS structural condition more faithfully. The overall process is illustrated in Fig.~\ref{fig:model}.

\subsection{Residual Denoising Formulation}

\begin{algorithm}[t]
\caption{Patch-wise Reverse Sampling with Gaussian Blending}
\label{alg:algorithm}
\small
\textbf{Input:} OBS image $\tilde{I}$; conditional networks $\epsilon_\theta(\cdot, \cdot, t)$ and $I_{\text{res},\theta}(\cdot, \cdot, t)$; Gaussian blending masks $\mathbf{P}_d$\\
\textbf{Output:} Preliminary translated modern character $I_0$
\begin{algorithmic}[1]
\STATE Sample initial noise $\mathbf{z}_T \sim \mathcal{N}(0, \mathbf{I})$
\STATE Initialize $I_T \gets \tilde{I} + \bar{\beta}_T \mathbf{z}_T$
\STATE Set stability constant $c \gets 10^{-8}$
\FOR{$t = T$ \textbf{downto} $1$}
    \STATE Initialize accumulators $\Sigma_t \gets \mathbf{0}$, $\Phi_t \gets \mathbf{0}$, and $\mathbf{M} \gets \mathbf{0}$
    \FOR{$d = 1$ \textbf{to} $8$}
        \STATE Extract local region $d$ from $I_t$ and embed it into a fixed-size canvas $I_t^{(d)}$
        \STATE Extract local region $d$ from $\tilde{I}$ and embed it into a fixed-size canvas $\tilde{I}^{(d)}$
        \STATE Predict noise $\hat{\epsilon}_t^{(d)} \gets \epsilon_\theta(I_t^{(d)}, \tilde{I}^{(d)}, t)$
        \STATE Predict residual $\hat{I}_{\text{res}, t}^{(d)} \gets I_{\text{res},\theta}(I_t^{(d)}, \tilde{I}^{(d)}, t)$
        \STATE Crop the valid local support from $\hat{\epsilon}_t^{(d)}$ and $\hat{I}_{\text{res}, t}^{(d)}$, and project it back to local region $d$
        \STATE $\Sigma_t \gets \Sigma_t + \operatorname{Proj}_d(\hat{\epsilon}_t^{(d)}) \odot \mathbf{P}_d$
        \STATE $\Phi_t \gets \Phi_t + \operatorname{Proj}_d(\hat{I}_{\text{res}, t}^{(d)}) \odot \mathbf{P}_d$
        \STATE $\mathbf{M} \gets \mathbf{M} + \mathbf{P}_d$
    \ENDFOR
    \STATE $\bar{\epsilon}_t \gets \Sigma_t / (\mathbf{M} + c)$
    \STATE $\bar{I}_{\text{res}, t} \gets \Phi_t / (\mathbf{M} + c)$
    \STATE Sample step noise $\mathbf{z}_t \sim \mathcal{N}(0, \mathbf{I})$ if $t > 1$, else $\mathbf{0}$
    \STATE $\text{res\_term} \gets (\bar{\alpha}_t - \bar{\alpha}_{t-1}) \cdot \bar{I}_{\text{res}, t}$
    \STATE $\text{noise\_term} \gets \left(\bar{\beta}_t - \sqrt{\bar{\beta}_{t-1}^2 - \sigma_t^2}\right) \cdot \bar{\epsilon}_t$
    \STATE $I_{t-1} \gets I_t - \text{res\_term} - \text{noise\_term} + \sigma_t \cdot \mathbf{z}_t$
\ENDFOR
\RETURN $I_0$
\end{algorithmic}
\end{algorithm}

Following RDDM~\cite{liu2024residual}, we redefine the diffusion process to transition from the target modern character image $I_0$ to a state heavily dependent on the conditional OBS image $\tilde{I}$. We define the domain residual as $I_{\text{res}} = \tilde{I} - I_0$.

\textbf{Forward Diffusion Process:} The forward process gradually adds noise and shifts the mean toward $\tilde{I}$. The marginal distribution at step $t$ is formulated consistently using schedule parameters $\bar{\alpha}_t$ and $\bar{\beta}_t$:
\begin{equation}
q(I_t \mid I_0, I_{\text{res}}) = \mathcal{N}(I_t; I_0 + \bar{\alpha}_t I_{\text{res}}, \bar{\beta}_t^2 \mathbf{I}),
\end{equation}
where $\mathbf{I}$ denotes the identity matrix. The parameter $\bar{\alpha}_t \in [0, 1]$ monotonically increases to 1 at $t = T$, and $\bar{\beta}_t$ represents the noise scale. Using the reparameterization trick, we can express $I_t$ as
\begin{equation}
I_t = I_0 + \bar{\alpha}_t I_{\text{res}} + \bar{\beta}_t \epsilon, \quad \epsilon \sim \mathcal{N}(0, \mathbf{I}).
\end{equation}
Notice that when $t = T$, $\bar{\alpha}_T \approx 1$, leading to the boundary condition $I_T \approx \tilde{I} + \bar{\beta}_T \epsilon$. This ensures that the reverse process initiates from a noisy version of the OBS image rather than pure Gaussian noise.

\textbf{Reverse Generation Process:} To reverse the process from $I_T$ to $I_0$, the model employs two coupled networks to predict the residual and noise simultaneously: $I_{\text{res},\theta}(I_t, \tilde{I}, t)$ and $\epsilon_{\theta}(I_t, \tilde{I}, t)$. The target image is estimated at each step as
\begin{equation}
I_{0,\theta} = I_t - \bar{\alpha}_t I_{\text{res},\theta} - \bar{\beta}_t \epsilon_{\theta}.
\end{equation}
The reverse transition step $p_\theta(I_{t-1} \mid I_t, \tilde{I})$ is parameterized as $\mathcal{N}(I_{t-1}; \mu_\theta, \sigma_t^2 \mathbf{I})$. Deriving from the posterior distribution $q(I_{t-1} \mid I_t, I_0, I_{\text{res}})$ and substituting our estimates, the mean $\mu_\theta$ is computed. The final sampling step becomes
\begin{equation}
\label{eq:reverse_step}
I_{t-1} = I_t - (\bar{\alpha}_t - \bar{\alpha}_{t-1}) I_{\text{res},\theta} - \left(\bar{\beta}_t - \sqrt{\bar{\beta}_{t-1}^2 - \sigma_t^2}\right) \epsilon_{\theta} + \sigma_t \mathbf{z},
\end{equation}
where $\mathbf{z} \sim \mathcal{N}(0, \mathbf{I})$, and $\sigma_t^2$ is the reverse step variance controlled by a stochasticity parameter $\eta \in [0, 1]$. Following the RDDM sampling schedule, $\sigma_t$ is chosen such that $\bar{\beta}_{t-1}^2 - \sigma_t^2 \geq 0$, which keeps the square-root term well-defined.

The network is optimized using a weighted combination of the residual loss and noise loss:
\begin{align}
L(\theta) &= L_{\text{res}}(\theta) + \lambda L_{\epsilon}(\theta), \\
L_{\text{res}}(\theta) &= \mathbb{E}_{t,I_0,\tilde{I},\epsilon} \left[\left\| I_{\text{res}} - I_{\text{res},\theta}(I_t, \tilde{I}, t) \right\|_2^2\right], \\
L_{\epsilon}(\theta) &= \mathbb{E}_{t,I_0,\tilde{I},\epsilon} \left[\left\| \epsilon - \epsilon_\theta(I_t, \tilde{I}, t) \right\|_2^2\right].
\end{align}

\subsection{Patch-wise Training Loss Adaptation}

As discussed in Sect.~\ref{sec:segmentation}, the paired OBS and target images are segmented into $D$ local patches $\tilde{I}^{(d)}$ and $I_0^{(d)}$ under the LSS strategy. The training objective averages prediction errors across patches and time steps. Accordingly, the expected loss functions are computed in a patch-wise manner:
\begin{align}
\hat{L}(\theta) &= \hat{L}_{\text{res}}(\theta) + \lambda\hat{L}_{\epsilon}(\theta), \\
\hat{L}_{\text{res}}(\theta) &= \mathbb{E}_{t,d,I_0,\tilde{I},\epsilon} \left[\left\| I_{\text{res}}^{(d)} - I_{\text{res},\theta} \big(I_t^{(d)}, \tilde{I}^{(d)}, t\big) \right\|_2^2\right], \\
\hat{L}_{\epsilon}(\theta) &= \mathbb{E}_{t,d,I_0,\tilde{I},\epsilon} \left[\left\| \epsilon^{(d)} - \epsilon_\theta \big(I_t^{(d)}, \tilde{I}^{(d)}, t\big) \right\|_2^2\right].
\end{align}
The goal is to minimize the difference between predicted and ground truth patch residuals $I_{\text{res}}^{(d)}$ and noise $\epsilon^{(d)}$. Averaging the prediction errors across patches helps the model learn the localized distribution of structural correspondences, preventing overfitting to individual patches. Here, each patch residual is defined as $I_{\text{res}}^{(d)}=\tilde{I}^{(d)}-I_0^{(d)}$, consistent with the global residual definition.

For pairs that do not meet the gating threshold, no spatial decomposition is applied and $D$ is set to 1. This allows the same loss formulation to cover both patch-wise and holistic training samples.

\subsection{Inference Process}

At inference time, since paired modern targets are unavailable, the LightGlue-based gate is not used. FROD adopts a fixed patch-wise reverse sampling strategy for every OBS input and reconstructs the final output using Gaussian blending. Specifically, the input OBS image $\tilde{I}$ is decomposed into eight overlapping local regions, and each local region is translated at the same $100 \times 100$ network resolution as holistic processing. This inference-time patch decomposition does not impose additional paired local correspondence supervision; it serves as an input decomposition and blending strategy for recovering local strokes while preserving global consistency. Algorithm~\ref{alg:algorithm} summarizes the patch-wise reverse sampling procedure. During training, gated samples use the patch-wise losses in Sect.~\ref{sec:segmentation}, whereas low-match samples use the same loss with $D=1$ as holistic supervision.

\section{Font Stylization Refinement}

\begin{figure}[t]
\centering
\includegraphics[width=\textwidth]{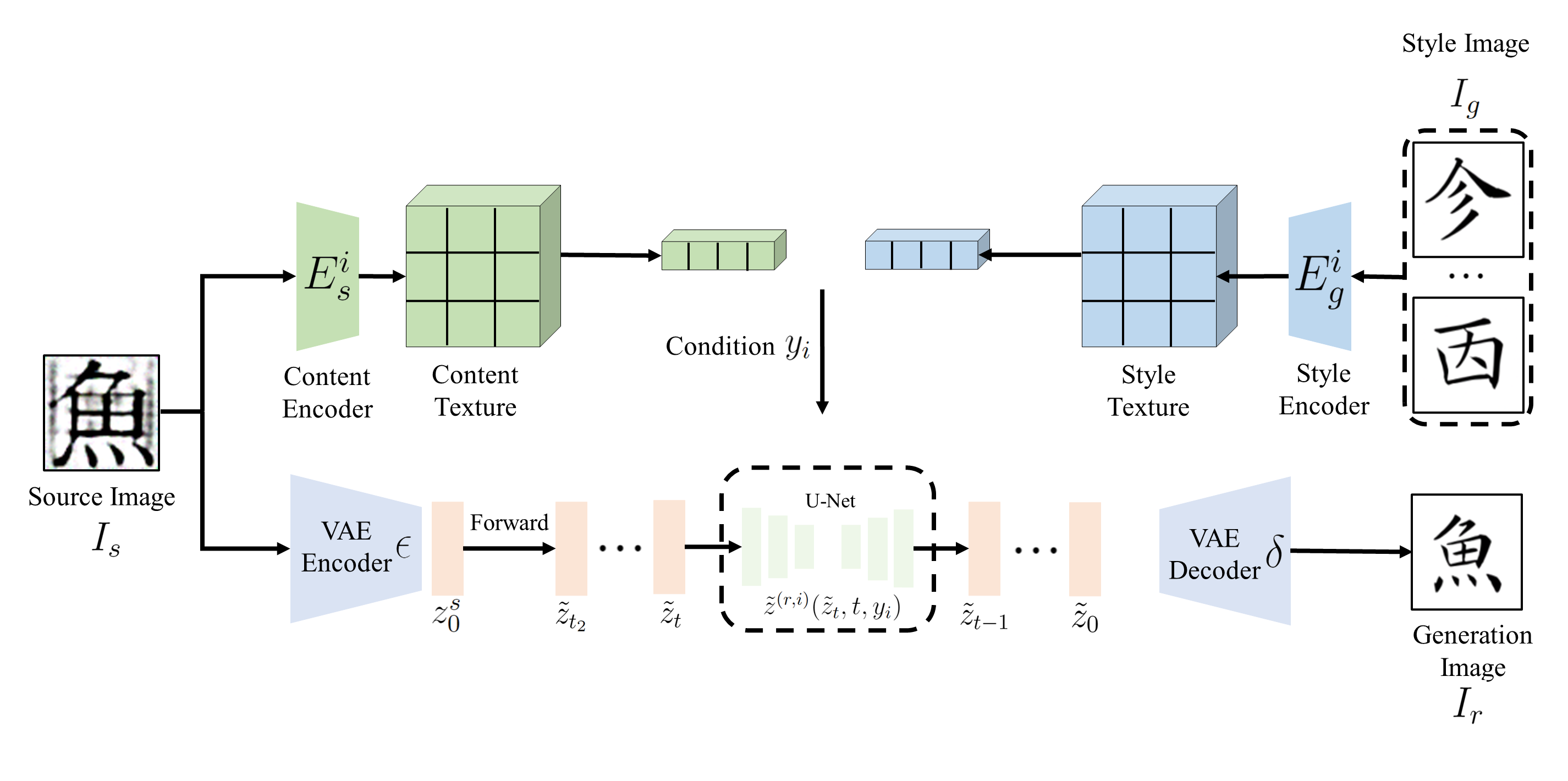}
\caption{Steps of font stylization refinement.}
\label{fig:font}
\end{figure}

Although RDDM substantially improves the recovery of modern character topology from oracle bone script, the preliminary outputs often retain raw brush textures, structural fragmentation, and artifacts inherited from the ancient domain. To bridge the large domain gap between these coarse outputs and standard printed modern fonts, which is crucial for reliable downstream OCR rather than mere visual refinement, we introduce a multi-stage font stylization refinement network. This stage preserves the core character structure while regularizing stroke widths and removing raster artifacts.

Inspired by few-shot font generation methods, we adopt the MSD-Font architecture proposed in~\cite{fu2024generate}. The font stylization refinement module maps coarse translated characters to clean, stylized modern fonts and thereby unifies visual representations for fair evaluation.

Although the modern font corpus covers the evaluated modern character categories, it contains only standard printed glyphs and no OBS images or OBS-modern paired samples. The font stylization refinement module is therefore trained independently as a modern-font structural and stylistic prior, rather than as OBS-to-modern paired supervision. During inference, the style image only provides font-level appearance guidance and is not the paired target glyph of the input OBS sample. This module serves as a post-generation standardization stage that regularizes the coarse RDDM output into a clean modern glyph form.

The overall MSD-Font architecture is shown in Fig.~\ref{fig:font}. We use the preliminary translation from the RDDM stage as the source image $I_s$. The source image is encoded by a Vector Quantized VAE (VQ-VAE) encoder into the latent feature $\tilde{z}_0^s$. In parallel, a style image $I_g$ is processed by the style encoder to produce the style condition $y_i$. The style image provides font-level appearance guidance and is not used as a paired target glyph for the test OBS input. During the forward diffusion process, Gaussian noise is added to $\tilde{z}_0^s$ to obtain the intermediate latent state $\tilde{z}_{t_2}$. During reverse diffusion, MSD-Font employs a style-conditioned prediction network $\tilde{z}^{(r,i)}(\tilde{z}_t, t, y_i)$ to progressively transform the source latent toward a clean stylized result, and the VQ-VAE decoder finally outputs the generated modern glyph $I_r$.

The reverse process consists of three distinct stages: glyph construction, font transformation, and refinement. Starting from the noisy latent state associated with the source image, the model first reconstructs the coarse content structure of $I_s$ during the glyph construction stage to anchor the basic topology. This stage simplifies to a single-step forward transition at an intermediate timestep $t_2$:
\begin{align}
\tilde{z}_{t_2} = \sqrt{\bar{\alpha}_{t_2}} \tilde{z}_0^s + \sqrt{1 - \bar{\alpha}_{t_2}} \epsilon_0,
\end{align}
where $\bar{\alpha}_{t_2}$ is the predefined cumulative diffusion coefficient at time $t_2$, and $\epsilon_0 \sim \mathcal{N}(0, \mathbf{I})$.

After obtaining the first-stage latent noise map $\tilde{z}_{t_2}$, the font transformation stage progressively transforms it into the intermediate stylized latent $\tilde{z}_{t_1}$ using a style-conditioned network $\tilde{z}^{(r,1)}(\tilde{z}_t, t, y_1)$. Finally, the font refinement stage uses a secondary conditional network $\tilde{z}^{(r,2)}(\tilde{z}_t, t, y_2)$ to refine $\tilde{z}_{t_1}$, repairing local stroke intersections and yielding the final latent representation for the generated character. The VQ-VAE decoder then produces the final clear, stylized modern Chinese character image $I_r$.

\section{Experimental Results and Analysis}

\subsection{Experimental Settings}

During training, the diffusion model is optimized with an initial learning rate of $2 \times 10^{-4}$. We maintain an Exponential Moving Average (EMA) of model parameters with a decay rate of 0.995 to improve optimization stability. Training is conducted for 350 epochs with a batch size of 16. All network inputs are resized to $100 \times 100$ pixels for both patch-wise and holistic processing. For gated samples, we use eight overlapping local regions with $p=64$ as the segmentation support, and the segmentation threshold is set to $\gamma=15$.

\subsection{Datasets and Evaluation Metrics}

We use OBC-V~\cite{zhou2025oracle} as the base dataset and augment it with supplementary OBS images from EVOBC~\cite{guan2024open} and HUST-OBC~\cite{wang2024open}. This augmentation is performed by adding real OBS samples from external datasets rather than synthesizing new glyph images. After duplicate removal, the augmented dataset contains 74,219 OBS images mapped to 1,590 interpreted modern Chinese character classes. To prevent class-level leakage and ensure rigorous evaluation, we implement a character-disjoint train-test split while maintaining an approximately 9:1 split ratio. Specifically, all OBS samples belonging to the same modern Chinese character class are assigned exclusively to either the training set or the test set, and no modern character class appears in both sets. We further remove duplicate OBS instances before splitting and ensure that instances extracted from the same OBS source image are not shared across splits.

We evaluate the framework using both image quality metrics and recognition accuracy. \textbf{Image Quality Metrics:} We report standard metrics, including FID, RMSE, SSIM, and LPIPS. The generated characters are compared directly with the corresponding standard printed modern Chinese fonts used as ground-truth references. \textbf{Recognition Accuracy (Top-k):} To evaluate isolated-glyph automatic decipherment objectively, we deploy an OCR model. Specifically, we train a ResNet-50 classifier on a standard corpus of printed modern Chinese characters covering all 1,590 evaluated classes. The OCR classifier is trained on standard printed modern Chinese fonts and is used solely as an automatic recognizer for generated modern glyphs. The generated translation outputs are fed into this classifier to compute Top-1, Top-5, Top-10, Top-100, and Top-500 accuracies.

For the OBSD~\cite{guan2024deciphering} baseline, we use the official implementation and retain its LSS-based initial decipherment and zero-shot refinement stages, so OBSD is evaluated as its complete published pipeline.

\subsection{Comparison and Analysis of Results}

Table~\ref{tab:quality} reports the quantitative image translation results. FROD achieves the best performance across all image-quality metrics. GAN-based methods, including Pix2Pix, CycleGAN, and DRIT++, perform poorly because they have limited capacity to model the severe structural deformations between OBS and modern characters. Diffusion-based methods generalize better, and FROD further improves over the strongest baseline, OBSD, by learning more robust structural priors through RDDM.

\begin{table}[t]
\caption{Comparative evaluation of image translation quality.}
\label{tab:quality}
\centering
\small
\begin{tabularx}{\textwidth}{lCCCC}
\toprule
Method & FID$\downarrow$ & RMSE$\downarrow$ & SSIM$\uparrow$ & LPIPS$\downarrow$ \\
\midrule
Pix2Pix & 246.08 & 0.5302 & 0.287 & 0.6037 \\
DRIT++ & 201.77 & 0.4877 & 0.313 & 0.5312 \\
CycleGAN & 212.46 & 0.4618 & 0.321 & 0.5133 \\
BBDM & 122.52 & 0.4181 & 0.458 & 0.3978 \\
OBSD & 45.18 & 0.3037 & 0.622 & 0.2226 \\
\textbf{FROD} & \textbf{35.92} & \textbf{0.2788} & \textbf{0.689} & \textbf{0.1798} \\
\bottomrule
\end{tabularx}
\end{table}

Table~\ref{tab:recognition} summarizes the OCR-based automatic decipherment results. FROD improves Top-1 accuracy over OBSD by 3.8 absolute percentage points (42.8\% vs. 39.0\%). This result indicates that the characters generated by FROD are more likely to be recognized as the correct modern character, suggesting that the proposed font stylization refinement and residual denoising mechanisms improve recognizability and complement the image-quality evaluation.

\begin{table}[t]
\caption{OCR-based automatic decipherment evaluation of generated images on the character-disjoint augmented dataset.}
\label{tab:recognition}
\centering
\small
\begin{tabularx}{\textwidth}{lCCCCC}
\toprule
Method & Top-1 & Top-5 & Top-10 & Top-100 & Top-500 \\
\midrule
Pix2Pix & 0.0\% & 0.0\% & 0.0\% & 1.9\% & 6.3\% \\
DRIT++ & 0.0\% & 0.0\% & 0.0\% & 2.5\% & 8.2\% \\
CycleGAN & 0.0\% & 0.0\% & 0.0\% & 13.8\% & 20.1\% \\
BBDM & 18.2\% & 22.0\% & 23.9\% & 33.3\% & 37.1\% \\
OBSD & 39.0\% & 42.1\% & 45.2\% & 57.9\% & 61.6\% \\
\textbf{FROD} & \textbf{42.8\%} & \textbf{44.0\%} & \textbf{47.2\%} & \textbf{59.1\%} & \textbf{62.9\%} \\
\bottomrule
\end{tabularx}
\end{table}

\begin{figure}[p]
\centering
\includegraphics[width=0.66\textwidth]{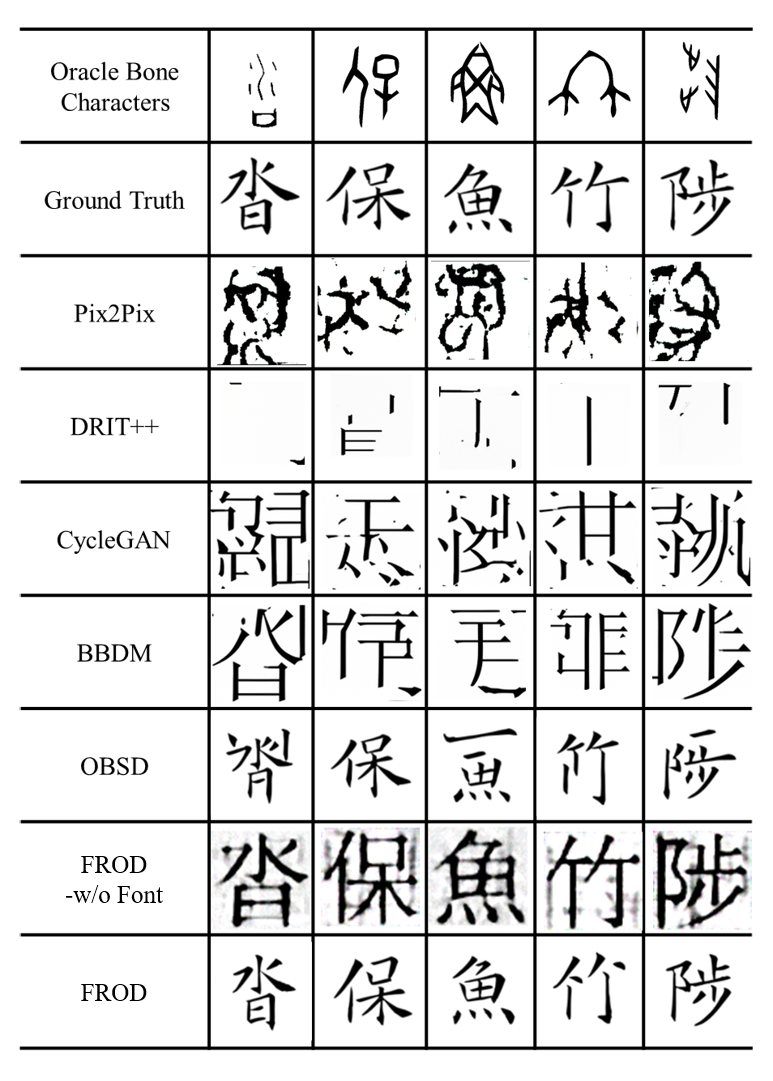}
\caption{Comparative analysis of oracle bone script image generation quality across different methods.}
\label{fig:result}
\end{figure}

Fig.~\ref{fig:result} presents a qualitative comparison. Although OBSD captures the overall structure, it frequently suffers from stroke omissions and misplacements because of its blind LSS strategy. The comparison between ``FROD-w/o Font'' and the full FROD output further shows that the font stylization refinement stage suppresses raster artifacts and regularizes stroke appearance. In contrast, FROD, trained with gated segmentation supervision and RDDM, reconstructs intricate stroke topologies with higher fidelity to modern character standards.

\begin{figure}[p]
\centering
\includegraphics[width=0.54\textwidth]{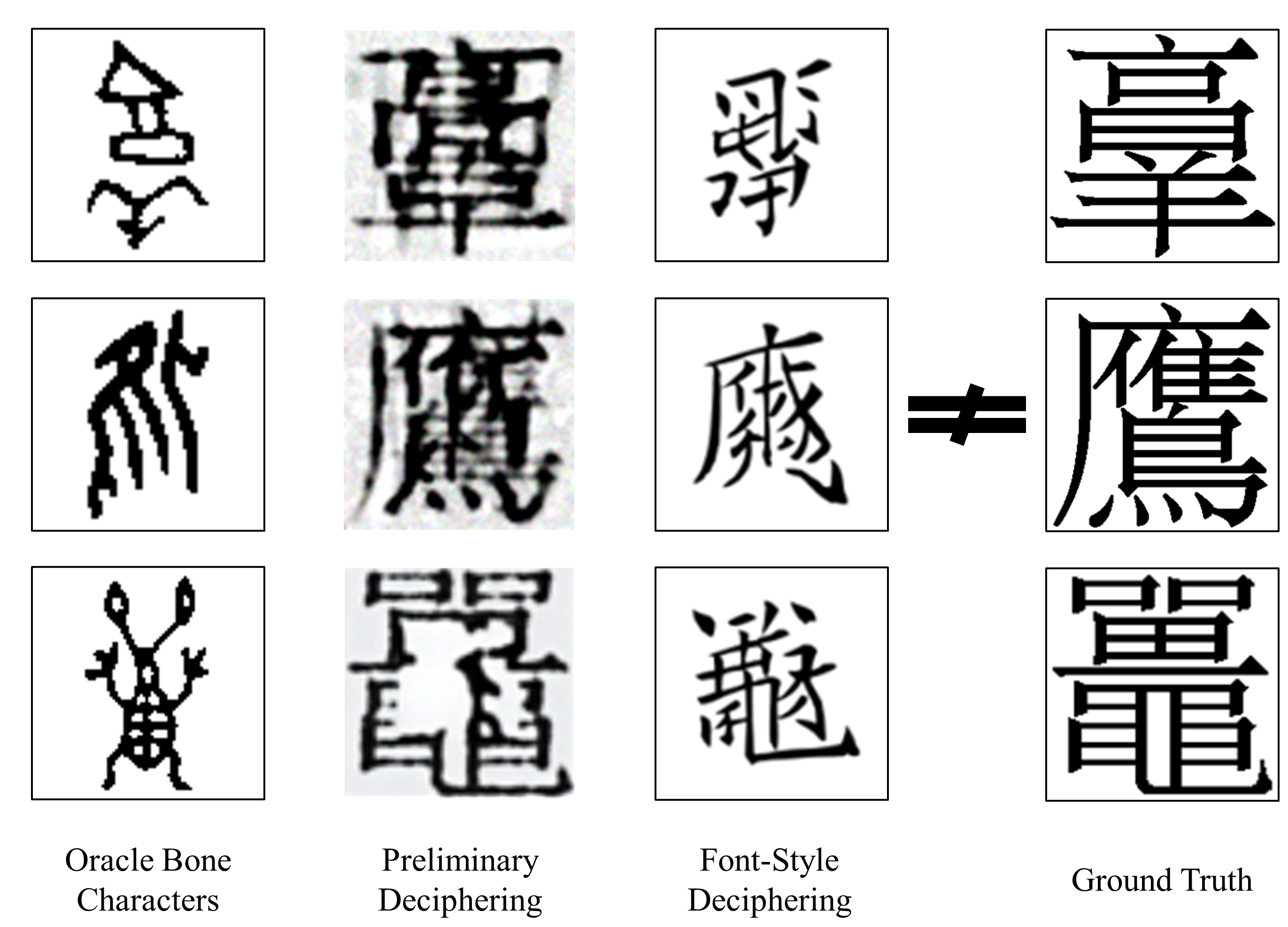}
\caption{Decipherment failure cases.}
\label{fig:failure}
\end{figure}

Fig.~\ref{fig:failure} also illustrates representative failure cases. The primary limitations arise from incomplete modeling of complex character structures and insufficient recovery of critical local details. In these examples, the preliminary translation preserves only coarse topology, and the stylization stage may amplify structural ambiguities rather than resolve them. These failures explain part of the remaining gap in both image-quality metrics and OCR accuracy.

\subsection{Ablation Study Results and Analysis}

To isolate and validate the contribution of each component, we conduct ablation studies in Table~\ref{tab:ablation}. We consider the following variants:
\begin{itemize}
\item \textbf{FROD-AlwaysSeg}: Removes the LightGlue threshold and blindly segments all training pairs.
\item \textbf{FROD-StandardDiff}: Replaces RDDM with a standard noise-predicting conditional diffusion model.
\item \textbf{FROD-w/o Font}: Removes the multi-stage font stylization refinement step.
\end{itemize}

\begin{table}[t]
\caption{Ablation study of FROD components.}
\label{tab:ablation}
\centering
\small
\begin{tabularx}{\textwidth}{lCCCC}
\toprule
Method & FID$\downarrow$ & RMSE$\downarrow$ & SSIM$\uparrow$ & LPIPS$\downarrow$ \\
\midrule
FROD-AlwaysSeg & 44.51 & 0.3121 & 0.601 & 0.2305 \\
FROD-StandardDiff & 49.33 & 0.3418 & 0.582 & 0.2511 \\
FROD-w/o Font & 56.99 & 0.3596 & 0.559 & 0.2833 \\
\textbf{FROD (Full)} & \textbf{35.92} & \textbf{0.2788} & \textbf{0.689} & \textbf{0.1798} \\
\bottomrule
\end{tabularx}
\end{table}

The results show that ``AlwaysSeg'' performs substantially worse than the gated strategy, which supports our hypothesis that blindly segmenting highly abstract OBS pairs damages structural integrity. Replacing RDDM with a standard diffusion model (``StandardDiff'') increases positional drift and degrades FID. Finally, removing the font stylization refinement stage (``w/o Font'') leaves noticeable edge noise and raster artifacts, further confirming that each module is important for producing high-quality translation results.

\subsection{Limitations}

Although the character-disjoint split evaluates generalization to unseen modern character classes while preventing shared source instances across training and testing, the reported OCR accuracy should still be interpreted as an automatic proxy for isolated-glyph decipherment rather than definitive philological decipherment. End-to-end decipherment ultimately requires contextual, philological, and archaeological evidence beyond isolated glyph images. In addition, the current evaluation reports aggregate performance over the full character-disjoint augmented test set; more fine-grained analysis by stroke complexity, radical composition, and structural deformation will be valuable in future dataset releases.

\section{Conclusion}

In this paper, we introduced FROD, an image translation framework for assisting the decipherment of oracle bone script by generating recognizable modern Chinese character candidates. To address the severe structural variation between ancient and modern scripts, FROD uses a LightGlue-based feature matching mechanism to provide gated segmentation supervision, thereby improving local radical alignment without forcing mismatched patch correspondences. At inference time, FROD adopts a fixed patch-wise reverse sampling strategy reconstructed with Gaussian blending. We further adapt RDDM to model residual signals explicitly, which mitigates the positional drift and stroke disorder prevalent in standard diffusion methods. Combined with a multi-stage font stylization refinement network, FROD produces clean and standardized modern character images. Extensive experiments show that the proposed method improves image generation quality and downstream OCR accuracy, yielding a 3.8\% absolute gain in Top-1 accuracy over OBSD on our character-disjoint augmented dataset.

In future work, we plan to explore vector-based font generation for synthesizing scalable outlines and to investigate the integration of multi-task perceptual recognition losses into diffusion training in order to further improve Top-1 accuracy and domain applicability.

\end{document}